\documentclass[10pt,twocolumn,letterpaper]{article}

\usepackage[pagenumbers]{cvpr} 

\usepackage{graphicx}
\usepackage{amsmath}
\usepackage{algorithm}
\usepackage{algpseudocode}
\usepackage{amssymb}
\usepackage{booktabs}
\usepackage{threeparttable}
\usepackage{adjustbox}
\usepackage{lipsum} 
\usepackage{multirow}
\usepackage[para]{footmisc}
\usepackage[accsupp]{axessibility}  
\makeatletter
\@namedef{ver@everyshi.sty}{}
\makeatother
\usepackage{hyperref}
\hypersetup{colorlinks, pagebackref, breaklinks}

\usepackage[capitalize]{cleveref}
\crefname{section}{Sec.}{Secs.}
\Crefname{section}{Section}{Sections}
\Crefname{table}{Table}{Tables}
\crefname{table}{Tab.}{Tabs.}

\def\cvprPaperID{8765} 
\def\confName{CVPR}
\def\confYear{2023}

\begin{document}

\setlength{\footnotesep}{0.1cm}
\setlength{\skip\footins}{0.5cm}

\title{SDC-UDA: Volumetric Unsupervised Domain Adaptation Framework for Slice-Direction Continuous Cross-Modality Medical Image Segmentation}


\author{ Hyungseob Shin$^{1*}$ \and Hyeongyu Kim$^{1*}$ \and Sewon Kim$^{4,5}$ \and Yohan Jun$^{7,8}$ \and Taejoon Eo$^{1,6}$  \and Dosik Hwang$^{1,2,3,9\dagger,}$ \\\normalsize \\ \normalsize \{$^{1}$School of Electrical and Electronic Engineering, \ $^{2}$Department of Oral and Maxillofacial Radiology, College of \\ \normalsize Dentistry, $^{3}$Department of Radiology and Center for Clinical Imaging Data Science, College of Medicine\} \\ \normalsize @Yonsei University \ \ $^{4}$Naver AI Lab \ \ $^{5}$Naver Cloud  \ \ $^{6}$Probe Medical, Inc. \ \ $^{7}$Martinos Center for Biomedical Imaging  \\ \normalsize $^{8}$Harvard Medical School \ \  $^{9}$Center for Healthcare Robotics, Korea Institute of Science and Technology \\ {\tt\small \{whatzupsup, lion4309, dosik.hwang\}@yonsei.ac.kr }}

\maketitle

\begin{abstract}
   Recent advances in deep learning-based medical image segmentation studies achieve nearly human-level performance  in fully supervised manner. However, acquiring pixel-level expert annotations is extremely expensive and laborious in medical imaging fields. Unsupervised domain adaptation (UDA) can alleviate this problem, which makes it possible to use annotated data in one imaging modality to train a network that can successfully perform segmentation on target imaging modality with no labels. In this work, we propose SDC-UDA, a simple yet effective volumetric UDA framework for \textbf{S}lice-\textbf{D}irection \textbf{C}ontinuous cross-modality medical image segmentation which combines intra- and inter-slice self-attentive image translation, uncertainty-constrained pseudo-label refinement, and volumetric self-training. Our method is distinguished from previous methods on UDA for medical image segmentation in that it can obtain continuous segmentation in the slice direction, thereby ensuring higher accuracy and potential in clinical practice. We validate SDC-UDA with multiple publicly available cross-modality medical image segmentation datasets and achieve state-of-the-art segmentation performance, not to mention the superior slice-direction continuity of prediction compared to previous studies. 
\end{abstract}

\renewcommand*{\thefootnote}{\fnsymbol{footnote}}
\footnotetext[1]{Equal contribution.} 
\renewcommand*{\thefootnote}{\fnsymbol{footnote}}
\footnotetext[2]{Corresponding author.} 

\section{Introduction}
\label{sec:intro}

\begin{figure}[t]
\captionsetup{width=1.0\linewidth}
\centering\includegraphics[width=1\linewidth]{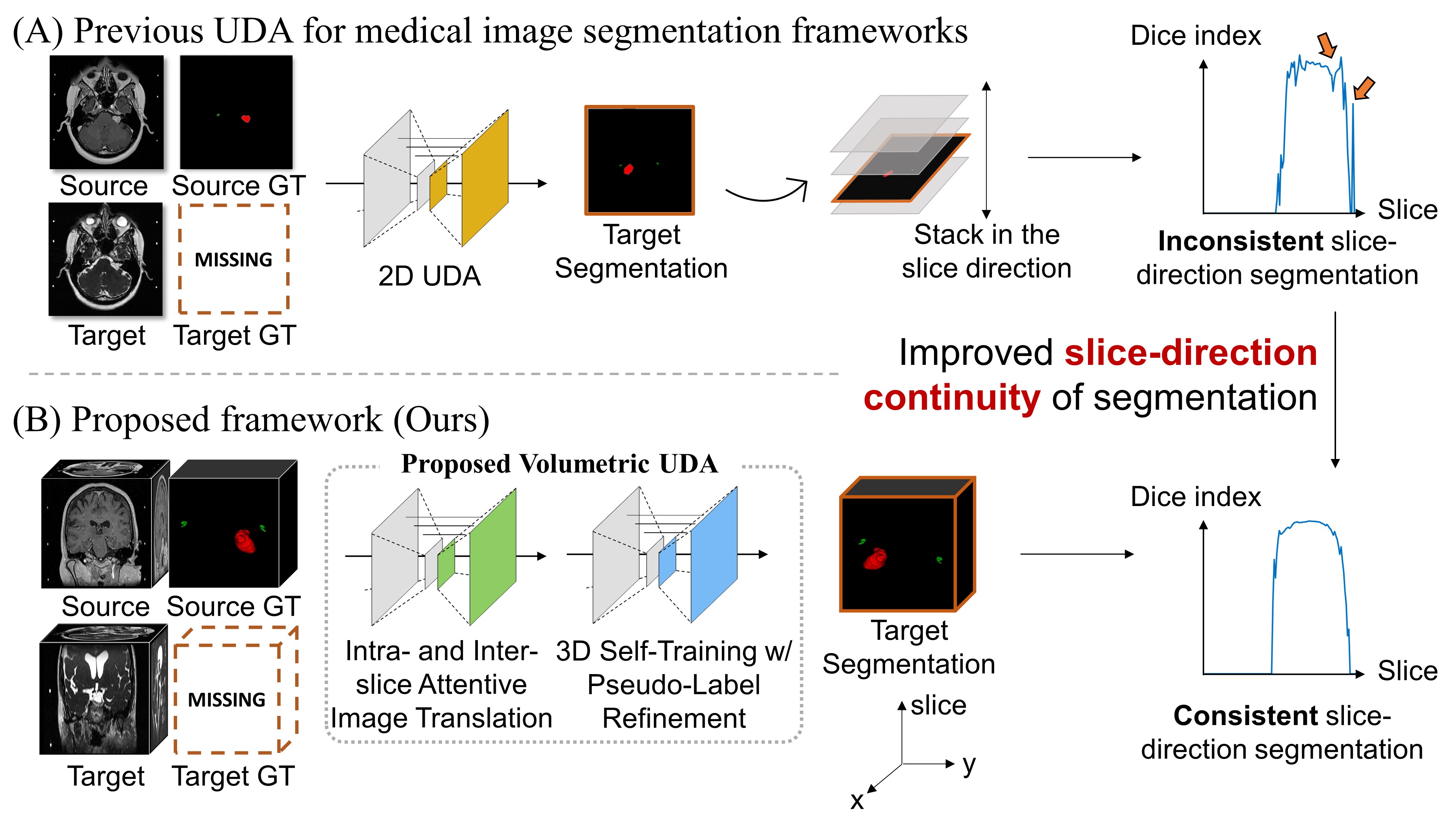}
\caption{ An illustration that describes the comparison between our proposed method with previous methods. (A) Previous UDA for medical image segmentation studies mostly utilize 2D UDA, which leads to inconsistent predictions in the slice direction when the predictions are stacked. (B) The proposed framework (\textbf{SDC-UDA}) considers volumetric information in the translation and segmentation process, respectively, which leads to improved slice-direction continuity of segmentation that is much practical for clinical use.}
\label{fig:fig1}
\end{figure}

\begin{figure*}[ht]
\captionsetup{width=1.0\linewidth}
\centering\includegraphics[width=1\linewidth]{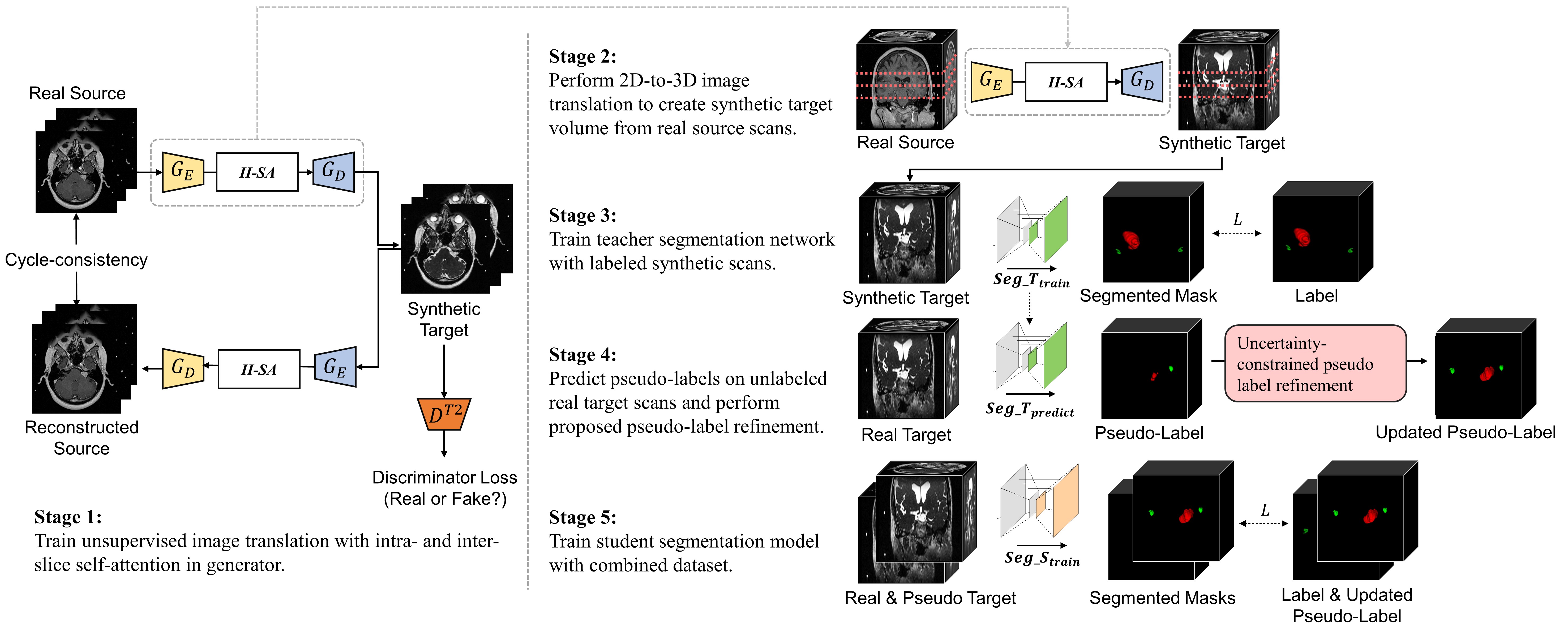}
\caption{Overview of our volumetric UDA framework. First, source-to-target image transformation is performed via unpaired image translation with intra- and inter-slice self-attention (stage 1-2). Second, volumetric self-training is performed (stage 3-5). During self-training, uncertainty-constrained pseudo-label refinement is conducted to improve pseudo-labels, thereby maximizing the effect of self-training (stage 4). The reverse loop of image translation is omitted for ease of illustration. Detailed architecture of image translation network is described in \cref{fig:fig25d}. Best viewed in color and on high-resolution display. \textit{II-SA}: Intra- and inter-slice self-attention.}
\label{fig:fig2}
\end{figure*}

With the surprising development of deep learning (DL), many studies are now showing remarkable performance in various applications\cite{eo2020accelerating, jun2021joint, lee2021relevance}. However, when a DL model faces data from an unseen domain, performance degradation occurs \cite{ganin2015unsupervised, tzeng2017adversarial}. Resolving this issue is important for the DL techniques to be applied in real world since collecting data from all domains and labeling them is very impractical and inefficient. Unsupervised Domain Adaptation (UDA) aims to alleviate this problem by adapting a model trained on source domain data to  target domain, without the necessity of supervision in the target domain.
Data dependency is more serious in medical image segmentation field since acquiring pixel-level expert annotation is extremely expensive and time-consuming \cite{chen2019synergistic, liu2019susan, zhang2018translating}. \\
Previous studies on UDA in the field of cross-modality medical image segmentation are normally conducted by simultaneously learning 2D image translation and target-domain segmentation, inferring the trained model on each slice of target data, and then stacking the predictions in the slice direction (\cref{fig:fig1}). As a result, they can lead to inconsistent or fluctuating predictions in the slice direction that are not revealed in quantitative metrics and may interrupt accurate analysis of target structure, making it difficult to use in real world clinical practice. In contrast, SDC-UDA can achieve \textbf{improved slice-direction continuity by incorporating volumetric natures of medical imaging}, which has not been thoroughly explored in previous works. It efficiently generates synthetic target volumes with neighbor-aware image translation by utilizing intra- and inter-slice self-attention module\cite{shin2022digestive}. Then, volumetric self-training is followed with uncertainty-constrained pseudo-label refinement strategy that adaptively increases the accuracy of pseudo-labels according to the target data (\cref{fig:fig2}). \textbf{Preliminary version of this work has won the 1st place in an unsupervised cross-modality domain adaptation for medical image segmentation challenge}\cite{DORENT2023102628, shin2022cosmos}. We updated the framework and extended it to multiple datasets. The main contribution of our work can be summarized as follows:

\begin{itemize}
    \item  We present \textbf{SDC-UDA}, a unified \textbf{volumetric UDA framework for cross-modality medical image segmentation}. 
    
    \item Intra- and inter-slice self-attention for efficient medical image translation: Proposed 2.5D translation framework with intra- and inter-slice self-attention module leads to \textbf{increased anatomy preservation and slice-direction smoothness in the synthesized volume}, enabling the synthetic volume to be used effectively in the following self-training steps.
    
    \item Volumetric self-training with uncertainty-constrained pseudo-label refinement: We propose a novel uncertainty-constrained pseudo-label refinement module that \textbf{can adaptively enhance the accuracy (i.e., sensitivity or specificity) of pseudo-labels}, thereby maximizing the performance of self-training on medical image segmentation.
    
    \item SDC-UDA was \textbf{validated on multiple public datasets with different data characteristics} for cross-modality medical image segmentation. It not only surpassed the performance of previous methods, but also showed \textbf{superior slice-direction segmentation continuity} which can provide precise analysis in clinical practice.
\end{itemize}
\section{Related Work}
\label{sec:related}

\subsection{UDA for medical image segmentation.}

UDA for semantic segmentation is one of the most popular themes among UDA-related studies \cite{hoffman2018cycada, kim2020learning, li2019bidirectional, tsai2018learning}. Especially, UDA on medical image segmentation is very attractive \cite{chen2019synergistic, Reuben_scirbble, huo2018synseg, jiang2020psigan, liu2019susan, zhang2018translating} since it can address the difficulty of obtaining expensive expert-level manual annotations.
Recent studies on UDA for medical image segmentation are mostly based on image adaptation in which domain translation and segmentation networks are trained end-to-end, thereby utilizing the source-to-target transformed images for training segmentation on target domain. \cite{huo2018synseg, liu2019susan, zhang2018translating} combined image-to-image translation and segmentation into a single network. \cite{chen2019synergistic} added feature adaptation with adversarial training using the segmentation output of the synthetic and real target domain images to better preserve the geometry of the structures-of-interest. Moreover, \cite{jiang2020psigan} added both the synthetic and real target domain images and their corresponding segmentation probability maps to adversarial training to preserve not only the geometry of target anatomies but also their appearance (i.e., intensities). \cite{han2021deep} utilized deep symmetric networks to further align the features of two domains. 
\\ \indent Our work differs from previous works in the following aspects: Previous works mostly train 2D-level image translation with target-domain segmentation and stack the segmentation of each target slice in the slice direction \cite{chen2020unsupervised, jiang2020psigan, tomar2021self, han2021deep}. Since the segmentation network does not consider the anatomical structure in the slice direction, this can lead to high variability of segmentation result even in adjacent slices. In contrast, SDC-UDA splits target-domain segmentation from image translation and therefore is not constrained to 2D framework. Consequently, SDC-UDA can effectively incorporate volumetric information in the translation and segmentation process, respectively. Moreover, in addition to the commonly used cross-modality medical image segmentation dataset \cite{zhuang2016multi, kavur2021chaos, landman2015miccai}, SDC-UDA has also been validated in the challenging task of cross-modality segmentation of small multi-class structures which are vestibular schwannoma (VS) and cochlea (i.e., CrossMoDA dataset\cite{DORENT2023102628, shapey2019artificial, shapey2021segmentation}). Previous studies locate target organs to the center of the preprocessed images and the majority of slices contain at least one target organ, whereas VS and cochleas in CrossMoDA dataset occupy extremely small fraction of the total voxels (0.028\% and 0.002\% for VS and cochlea, respectively, compared to approximately 3\% in average in another dataset\cite{zhuang2016multi}) which more reflects the real-world clinical imaging environment \cite{chen2020unsupervised, jiang2020psigan}. 

\subsection{Self-training on UDA for medical image segmentation.} 
Self-training belongs to semi-supervised learning which has emerged to improve the resources and cost put into data labeling \cite{xie2020self, zoph2020rethinking}. In self-training, a teacher model is first trained using only the labeled data. Next, pseudo-labels with high confidence are inferred by passing the unlabeled data on the trained model. With the labeled data and pseudo-labeled data, a larger dataset can be used to train a student model that performs better than the teacher model trained only on labeled data. Numerous attempts were made to apply self-training into semantic segmentation \cite{pastore2021closer, zhu2020improving, zou2018unsupervised, zou2020pseudoseg} as pixel-level annotations are expensive. Since pseudo-labels tend to be noisy labels, previous studies attempted to refine pseudo-labels to increase accuracy, thereby providing better supervision to unlabeled data. \cite{vs2022target} updated pseudo-labels by enhancing the sensitivity of pseudo-labels via selective voting. \cite{thompson2022pseudo} proposed region growing on the super-pixels where the pseudo-labels serve as the initial seed points. \cite{groger2021strudel} utilized uncertainty-weighted binary cross entropy loss to penalize high uncertainty region.  \\
\indent The proposed pseudo-label refinement strategy is effective in two aspects: 1) sensitivity and specificity-enhancing refinement can be adaptively used according to the characteristic of target data, and 2) it is a safer way to handle uncertainty maps since even the correctly segmented regions can have high uncertainty, and using uncertainty map to weight the loss function in the following training can therefore guide the model to wrong direction.     

\begin{figure*}[t]
\captionsetup{width=1.0\linewidth}
\centering\includegraphics[width=1\linewidth]{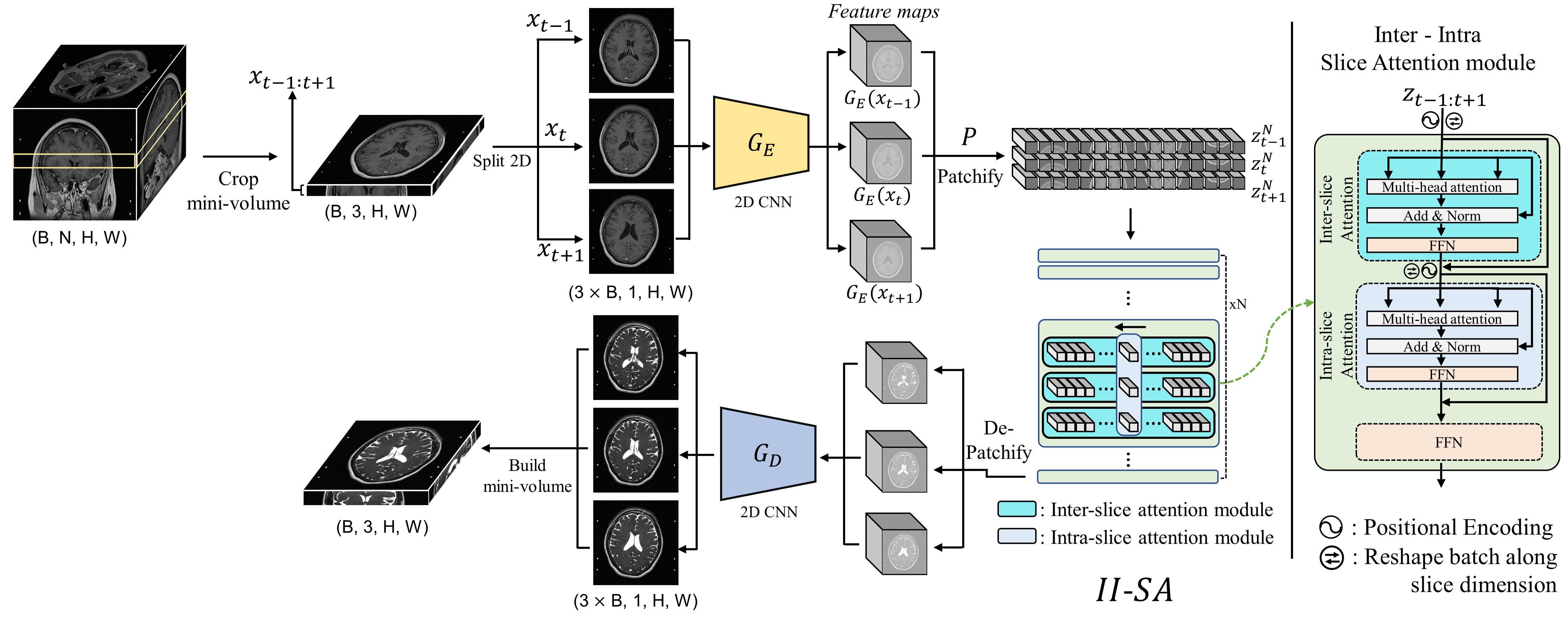}
\caption{Illustration of the proposed 2.5D image translation with intra- and inter-slice self-attention module. Source domain volume data is cropped to a mini-volume, and fed to CNN encoder $G_E$ as a stack of batches. The weights of $G_E$ are shared during batch update. The encoded feature maps are embedded to non-overlapping patches, which become the input of intra- \& inter- slice attention module $A$. Within $A$, intra- and inter-slice attention learns the volumetric information. }
\label{fig:fig25d}
\end{figure*}

\section{Methods}
\label{sec:methods}

\subsection{Unpaired image translation with intra- and inter-slice self-attention module.}

Many recent works on UDA for medical image segmentation have been developed on 2D framework, without considering the volumetric nature of medical imaging (2D multi-slice sequence or 3D sequence) \cite{chen2019synergistic, jiang2020psigan, huo2018synseg, liu2019susan}. Previous 2D UDA approaches split the 3D volume into 2D slices, and re-stack their translations into 3D volume afterward. Since the slices are processed individually, re-building the translated volume usually requires additional post-processing such as slice-direction interpolation, which still can not perfectly resolve problems such as slice-direction discontinuity. 
This remain a problem when conducting following steps such as self-training for volumetric segmentation. Although there exist frameworks such as 3D-CycleGAN which aim to conduct translation 3D volume-wise, they are rarely used due to typical drawbacks such as optimization complexity, computational burden, or the degradation of translation quality \cite{cciccek20163d, sun2022double}.



    

To remedy both lack of volumetric consideration of 2D and optimization-efficiency issue of 3D methods, we propose a simple and effective pixel-level domain translation method for medical volume data by translating a stack of source domain images into target domain with both intra- and inter-slice self-attention module. Unlike previous 2D methods that only translated within a single slice, our approach leverages information from adjacent slices in the slice-direction. This is similar to recent advances in video processing, which exploit information both within and between frames \cite{bertasius2021space, arnab2021vivit}.  In contrast to 3D methods that require expensive computational cost, ours do not necessitate heavy computation, enabling translation without intense down-sampling which 3D-based translations suffer from \cite{sun2022double}.

Shown in Fig.\ref{fig:fig25d}, proposed framework consists of sequential modules of encoder $G_E$, intra- and inter- slice self-attention \textbf{\textit{II-SA}} and decoder $G_D$. 
Taking three continuous slices ${x}_{t-1} , {x}_t, {x}_{t+1}$ of a volume as input, they are first fed to a 2D CNN encoder $G_E$ separately. Encoded features $ G_E({x}_{t-1}), G_E({x}_{t}), G_E({x}_{t+1}) $ are then fed to transformer-based attention module \textbf{\textit{II-SA}} together, as feature maps from continuous frames. Before $\textbf{\textit{II-SA}}$, each slice is embedded into small patches \textbf{z} of size $p$, as $ \textbf{z}_{t-1:t+1}^{n} = \textit{P}(G_{E}(x_{t-1:t+1})) \vert_{n=1:N}$, where $N = W_{G_{E}(x)} \cdot H_{G_{E}(x)}/p^{2}$. $W_{G_{E}(x)}, H_{G_{E}(y)}$ are each width, height of encoded feature map $G_{E}(x)$, respectively. In transformer-based attention module \textbf{\textit{II-SA}}, two types attention $A_{inter}$ and $A_{intra}$ exist. First, for the consideration of adjacent slices, learnable positional encoding is added to capture relative positional information within mini-volume patches. An inter-slice attention module is $A_{inter} =  A(\textbf{z}_{t-1}^{k}, \textbf{z}_{t}^{k}, \textbf{z}_{t+1}^{k})|_{k=1:N} $. And for the consideration of nearby pixels within a slice, an intra-slice self-attention exists, as  $A_{intra} =   A(\textbf{z}_{j}^{1},\textbf{z}_{j}^{2}, ... , \textbf{z}_{j}^{N})|_{j=t-1:t+1} $. The attention module is identical for both intra- and inter-slice attention, which is
\begin{equation}
 A = \sigma(\cfrac{QK^T} {\sqrt{d_K}})V
\label{eq:56}
\end{equation}
, where $Q=W_Q \cdot LN(\textbf{z})$, $K=W_K \cdot LN(\textbf{z})$, and $V=W_V \cdot LN(\textbf{z})$ are query, key and value embeddings, respectively. $LN$ denotes layer-normalization, $\sigma$ denotes soft-max operation and $d_K$ denotes embedding dimension.

For maximum efficiency of translation, inter-slice attention occurs on the adjacent patches over slice-direction only for computational efficiency. After attention module, a 2D CNN decoder $G_D$ reconstructs partial volume sets in the same way as done in $G_E$, constructing a translated 2.5D mini-volume. In inference step, only the center slices of mini-volume is stacked one by one to construct a whole translated volume.

\subsection{Volumetric self-training with uncertainty-constrained pseudo-label refinement.}
Even if the source-to-target transformed images are well generated, there may still exist some distribution gap with the real target domain data, and it is difficult to completely replace it. With self-training, pseudo-labels for the unlabeled target domain data can be obtained and used to provide supervision to the target domain data, thereby mitigating the domain gap from the model's perspective. We propose to utilize volumetric self-training to close the domain gap, with a simple and novel pseudo-label refinement strategy to maximize the effect of self-training. The steps are as follows. \\

\noindent \textbf{3.2.1. Training segmentation with labeled synthetic scans.} With the synthetic data $\tilde{x}^{t}$ converted from source domain and the annotations $y^s$ on the source scans (i.e., labeled synthetic dataset), we first train a teacher segmentation network $\textit{f}_{teacher}$ that minimizes the segmentation loss:

\begin{equation}
\mathcal{L} = \sum L_{seg}(y^s, f_{teacher}(\tilde{x}^t))
\label{eq:1}
\end{equation}

\begin{figure}[t]
\captionsetup{width=1.0\linewidth}
\centering\includegraphics[width=1\linewidth]{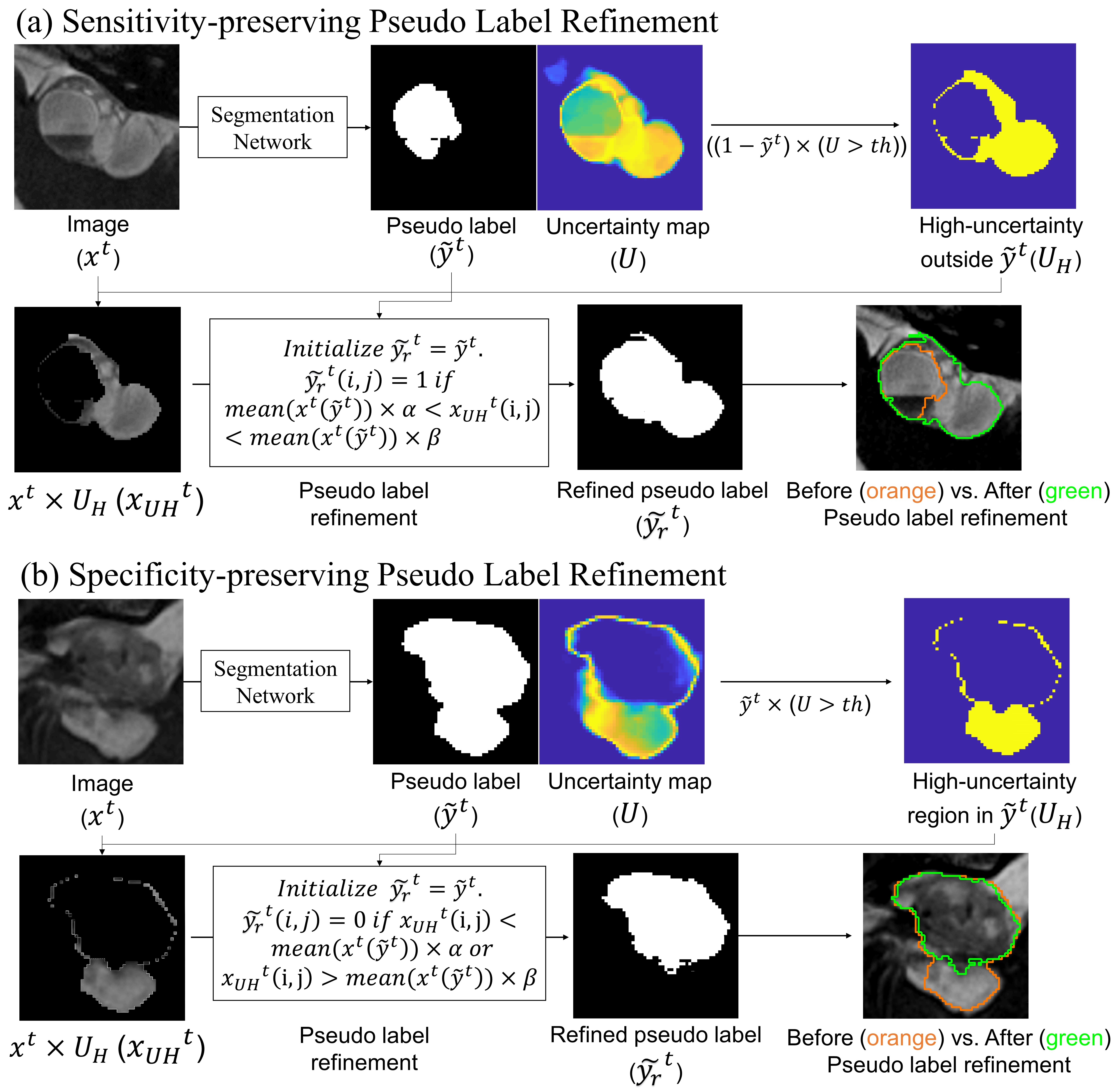}
\caption{Illustration of the proposed uncertainty-constrained pseudo-label refinement module on vestibular schwannoma of CrossMoDA dataset. Sensitivity/specificity enhancing refinement can be performed in parallel, and either one or a combination of the two can be used according to the characteristics of each class in the dataset. Best viewed in color and on high-resolution display.} 
\label{fig:fig4}
\end{figure}

\noindent \textbf{3.2.2. Inferring pseudo-labels on unlabeled target domain data.} Once the teacher model is trained, pseudo-labels $\tilde{y}^{t}$ of the non-annotated real data $x^{t}$ can be obtained by passing real target scans to the trained segmentation model $f_{teacher}$.

\begin{equation}
\tilde{y}^{t}_i = f_{teacher}( \{ x_i^t \}_{i=1}^{N_t} )
\label{eq:2}
\end{equation}

\noindent \textbf{3.2.3. Uncertainty-constrained pseudo-label refinement for sensitivity / specificity enhancement.} Since the pseudo-labels are noisy labels, they must be refined to increase accuracy and guide self-training to better direction. We devise a sensitivity and specificity enhancing pseudo-label refinement module that refines pseudo-labels based on image intensities, current pseudo-label, and high-uncertainty regions. \\

\noindent \textbf{Sensitivity-enhancing pseudo-label refinement.}
With the inference of pseudo-labels, uncertainty (i.e., entropy) maps corresponding to each class are computed according to the following equation:

\begin{equation}
\mathnormal{U} = \mathnormal{p}\log{\mathnormal{p}}
\label{eq:1}
\end{equation}

\noindent , where \emph{p} is the output probability map of each class. To enhance the sensitivity of pseudo-labels, highly uncertain regions outside the range of pseudo-labels are detected. Then, if the pixel intensity in this region is within a certain range of image intensity included in the current pseudo-label, this region is included to be part of pseudo-label. The equation can be formulated as:
\begin{equation}
\begin{split}
& \  \  \  \ \ \  \ \ \ \ \ \ \ \ \ \ \ \ \ \ \ \  \text{Initialize} \ \tilde{y}_r^t=\tilde{y}^t. \ \tilde{y}_r^t(i, j)=1 \text { if } \\
&\mathnormal{mean} \left(x^t\left(\tilde{y}^t\right)\right) \times \alpha<x_{U H}^t(i, j)<\mathnormal{mean}\left(x^t\left(\tilde{y}^t\right)\right) \times \beta
\label{eq:2}
\end{split}
\end{equation}
where $x^t$, $\tilde{y}^t$, $\tilde{y}_r^t$, and $x_{U H}^t$ represent target domain image, pseudo-label, refined pseudo-label, and image cropped with high-uncertainty region mask, respectively. This approach is grounded on the assumption that 1) pixels that have similar intensity and 2) are close to each other in medical image are likely to belong to the same class. The workflow is described with images in \cref{fig:fig4}. \\ \\
\noindent \textbf{Specificity-enhancing pseudo-label refinement.}
To enhance specificity of pseudo-labels, highly uncertain regions inside the range of pseudo-labels are detected. Then, if the pixel intensity in this region is outside a certain range of image intensity included in the current pseudo-label, it is excluded from the current pseudo-label. 
\begin{equation}
\begin{aligned}
\text{Initialize} \ \tilde{y}_r^t=\tilde{y}^t. \ \tilde{y}_r^t(i, j)=0 \text { if } \\ x_{U H}^t(i, j)<\mathnormal{mean} \left(x^t\left(\tilde{y}^t\right)\right) \times \alpha \ or \\ \ x_{U H}^t(i, j)> \mathnormal{mean}\left(x^t\left(\tilde{y}^t\right)\right) \times \beta
\end{aligned}
\label{eq:2}
\end{equation}                                               

\noindent \textbf{3.2.4. Retraining segmentation with combined data.} Synthetic target scans have distribution gap with the real target scans, but they are paired with perfect annotations. On the other hand, real target scans are paired with pseudo-labels, which are yet incomplete. We incorporate both pairs to self-training, to maximize the generalization ability and minimize the performance degradation caused by difference in distributions. With the combined data of labeled synthetic target scans $(\tilde{x}^t, y^s)$ and pseudo-labeled real target scans $({x}^t, \tilde{y}_r^t)$, we train a student segmentation $\textit{f}_{student}$ to minimize 
\begin{equation}
L= \sum L_{seg}(y^s, f_{student}(\tilde{x}^t)) 
+ 
\\ \sum L_{seg}(\tilde{y}_r^{t}, f_{student}(x^t))
\label{eq:8}
\end{equation}
Despite not being ground truth labels, it has been reported in the previous literature that the self-training scheme and the pseudo-labels increases the performance and generalization ability of the model on unseen data by utilizing unlabeled data into training \cite{xie2020self}.

\section{Experiments}
\label{sec:exp}

\subsection{Dataset}

\noindent \textbf{CrossMoDA dataset. } 
CrossMoDA dataset \cite{DORENT2023102628} for vestibular schwannoma and cochlea segmentation consists of 105 labeled contrast-enhanced T1 (ceT1) MRI scans and 105 unlabeled high-resolution T2 (hrT2) MRI scans. The direction of domain adaptation was from ceT1 to hrT2 since the annotations of hrT2 scans are not publicly available. Evaluation was performed online on 32 inaccessible hrT2 scans through an official leaderboard. Please refer to the supplementary material for examples of each domain data. \\

\noindent \textbf{Cardiac structure segmentation dataset.}
2017 Multi-Modality Whole Heart Segmentation (MMWHS) challenge dataset \cite{zhuang2016multi} which consists of 20 MRI and 20 CT volumes was used for cardiac segmentation. Domain adaptation was performed from MRI to CT, with 80\% of each modality used for training and 20\% (i.e., 4 volumes) of randomly selected CT scans used for evaluation. Target cardiac structures for segmentation were ascending aorta (AA), left atrium blood cavity (LAC), the left ventricle blood cavity (LVC), and myocardium of left ventricle (MYO). With a fixed coronal plane resolution of 256$\times$256, MRI and CT volumes were manually cropped to cover the 4 cardiac substructures. Please refer to the supplementary material for examples of each domain data.\\

\vspace*{-2mm}
\subsection{Implementation details}

\noindent \textbf{Unpaired image translation with intra- and inter-slice self-attention module.} 
Intra- and inter-slice attentive translation is based on 2D framework, where image slices are fed to 2D CNN \textit{E} as a stacked batches of size $(B\times3, 1, H, W)$, where \textit{H} and \textit{W} each denote height and width. After $G_E$, encoded feature map is reshaped to $(B, 3, H, W)$ and fed to the attention module for the consideration of slice-direction attention. Patch size \textit{p} was set as 2. Transposed convolution is used to convert embedded patches back to 2D feature maps. For a stack of mini-volume generated at $G$, we implement 2D discriminator $D$ for adversarial training of domain translation. For real input to $D$, 3 consecutive slices were randomly extracted from the target volumes. \\

\noindent \textbf{Uncertainty-constrained pseudo-label refinement. } 
For each class, grid searching was conducted on how to set the threshold for masking the high-uncertainty region and also on whether to use either one, or both of the sensitivity and specificity enhancing refinement. For CrossMoDA dataset, combining sensitivity and specificity-enhancing refinement in both classes (i.e., VS and Cochlea) with uncertainty threshold of 0.3 led to the best result. For cardiac dataset, the same threshold value was used and using only specificity-enhancing refinement for all classes except one (i.e., AA) led to the best result. This is attributed to the fact that the contrast difference between adjacent substructures was very weak in cardiac CT which led to noisier pseudo-label when sensitivity-enhancing module was used. Please refer to the tables in supplementary material for ablation results. $\alpha$ and $\beta$ were set as 0.6 and 1.4, respectively.  \\

\begin{table}[t]
\caption{Ablation study on the components of the proposed method on CrossMoDA dataset. ST and PL denote self-training and pseudo-label, respectively. Best results are bolded.}
\begin{tabular}{rcccc}
\multicolumn{5}{c}{CrossMoDA (T1$\rightarrow$T2)} \\ \hline
\multicolumn{1}{l}{} & \multicolumn{2}{c|}{Dice ($\uparrow$)} & \multicolumn{2}{c}{ASSD ($\downarrow$)} \\ \hline
\multicolumn{1}{l|}{Methods} & VS & \multicolumn{1}{l|}{C} & VS & C \\ \hline
\multicolumn{1}{l|}{Baseline} & 71.5 &  \multicolumn{1}{l|}{72.7} & 5.91 & 1.83 \\
\multicolumn{1}{l|}{+ intra/inter-slice attention} & 80.3 &  \multicolumn{1}{l|}{74.5} & 1.66 & 0.67 \\
\multicolumn{1}{l|}{+ ST w/o PL refinement} & 83.2 &  \multicolumn{1}{l|}{76.7} & 0.58 & 0.79 \\
\multicolumn{1}{l|}{+ ST w/ PL refinement} & \textbf{84.6} &  \multicolumn{1}{l|}{\textbf{84.9}} & \textbf{0.51} & \textbf{0.14}
\end{tabular}
\label{tab:table1}
\end{table}

\begin{figure}[t]
\captionsetup{width=1.0\linewidth}
\centering\includegraphics[width=1\linewidth]{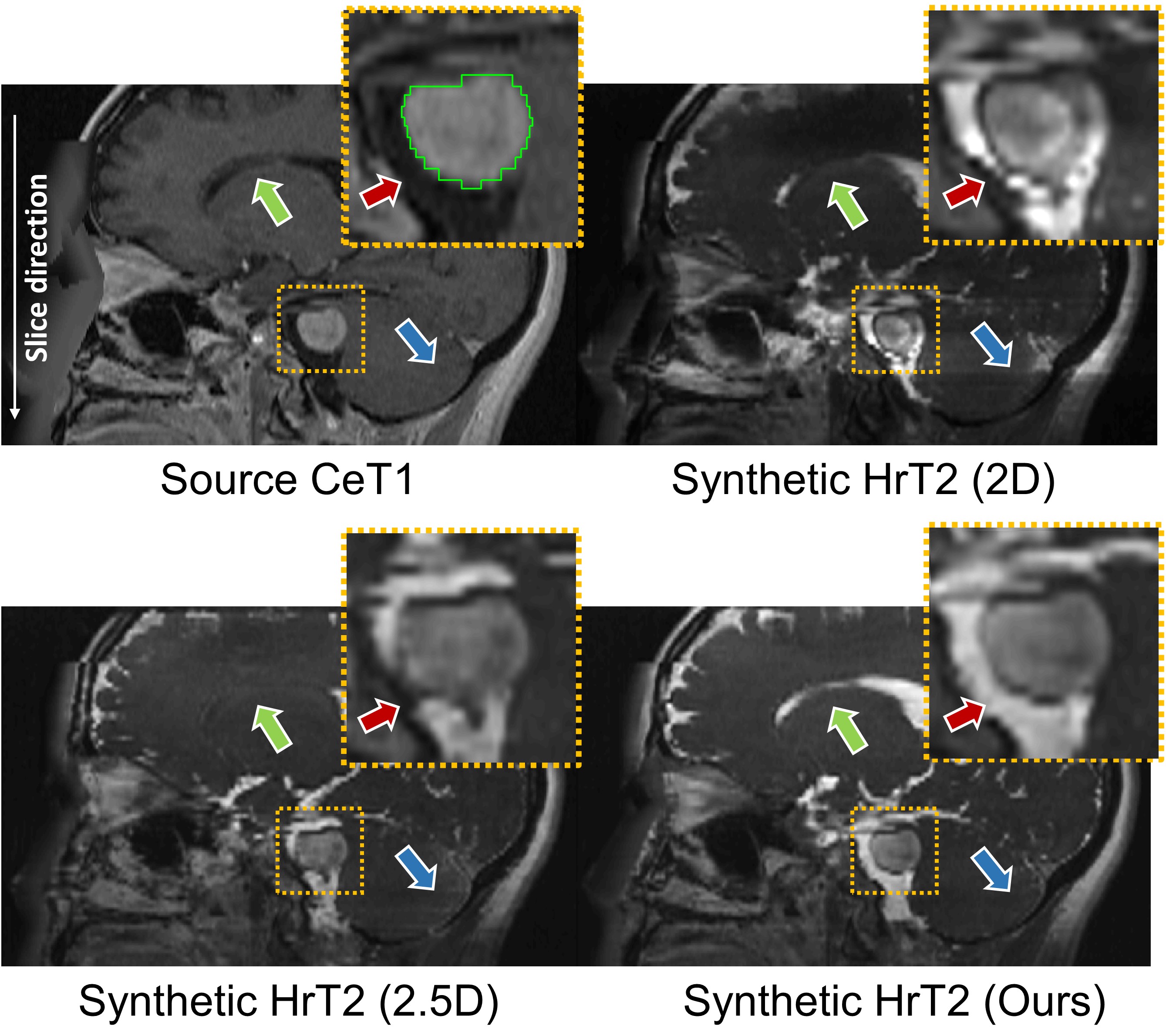}
\caption{Representative case showing the effect of intra- and inter-slice self-attentive image translation. Translation was performed using axial slices and the resulting volume is being observed from sagittal direction. Best viewed in color and on high-resolution display. }
\label{fig:fig5}
\end{figure}

\begin{table*}[t]

\caption{Comparison of quantitative results between SDC-UDA and previous non-medical and medical UDA methods. CycleGAN, CyCADA, ADVENT, and FDA represent non-medical UDA methods while SIFA and PSIGAN represent recent medical UDA methods. Best results are bolded.}
\centering
\small

\begin{adjustbox}{width = \textwidth}
\begin{threeparttable}
{

\begin{tabular}{ c ll lll  lll  lllll  lllll}
\multicolumn{2}{l}{} & \multicolumn{6}{c}{\textbf{VS and Cochlea (T1 $\rightarrow$ T2)}}  & \multicolumn{10}{c}{\textbf{Cardiac structures (MR $\rightarrow$ CT)}}\\ \cline{3-7} \cline{8-18} 
\multicolumn{2}{l|}{} & \multicolumn{3}{|c|}{\textbf{Dice} ($\uparrow$) } & \multicolumn{3}{c|}{\textbf{ASSD} ($\downarrow$)} & \multicolumn{5}{c|}{\textbf{Dice} ($\uparrow$) } & \multicolumn{5}{c}{\textbf{ASSD} ($\downarrow$)} \\ \cline{3-7} \cline{8-18} 

 \multicolumn{1}{c}{} & \multicolumn{1}{c|}{\textbf{Methods}} & \multicolumn{1}{|c}{VS} & \multicolumn{1}{c}{C} & \multicolumn{1}{c|}{Mean} & \multicolumn{1}{c}{VS} & \multicolumn{1}{c}{C} & \multicolumn{1}{c|}{Mean} & \multicolumn{1}{c}{AA} & \multicolumn{1}{c}{LAC} & \multicolumn{1}{c}{LVC} & \multicolumn{1}{c}{MYO} & \multicolumn{1}{c|}{Mean} & \multicolumn{1}{c}{AA} & \multicolumn{1}{c}{LAC} & \multicolumn{1}{c}{LVC} & \multicolumn{1}{c}{MYO} & \multicolumn{1}{c}{Mean} \\ \cline{2-18}
 
\multicolumn{1}{l|}{} &\multicolumn{1}{l|}{Fully-supervised \tnote{1}} &92.5 &87.7 &\multicolumn{1}{l|}{90.1} &0.2  &0.1  &\multicolumn{1}{l|}{0.15} &95.9 &92.0 &93.0 &88.2 &\multicolumn{1}{l|}{92.3} &1.0 &2.5 &1.8 &1.7 &1.8 \\
\multicolumn{1}{l|}{} &\multicolumn{1}{l|}{w/o adaptation}   &11.5 &0.0  &\multicolumn{1}{l|}{5.8}  &24.0 &NA   &\multicolumn{1}{l|}{NA}   &48.4 &52.0 &20.0 &4.7  &\multicolumn{1}{l|}{31.3} &18.8 &34.5 &32.0 &36.4 &30.4 \\ \hline

\multicolumn{1}{l|}{\multirow{4}{*}{\rotatebox[origin=c]{90}{\scriptsize Non-Medical}}} &\multicolumn{1}{l|}{CycleGAN \cite{zhu2017unpaired}}         &39.7 &0.0   &\multicolumn{1}{l|}{19.8}   &10.1 &NA   &\multicolumn{1}{l|}{NA}   &58.3 &59.8 &61.5 &23.1 &\multicolumn{1}{l|}{50.7} &11.4 &10.6 &9.2 &14.8 &11.5 \\
\multicolumn{1}{l|}{}             &\multicolumn{1}{l|}{CyCADA \cite{hoffman2018cycada}}           &39.2 &0.1  &\multicolumn{1}{l|}{19.7} &10.5 &18.1 &\multicolumn{1}{l|}{14.3} &57.2 &62.1 &65.0 &47.2 &\multicolumn{1}{l|}{57.9} &8.4 &9.0 &8.2 &7.9 &8.4 \\
\multicolumn{1}{l|}{}           & \multicolumn{1}{l|}{ADVENT \cite{vu2019advent}}           &8.3    &0.0  &\multicolumn{1}{l|}{ 4.2 }  & 17.1   & NA   &\multicolumn{1}{l|}{ NA }  &68.2 &71.8 &78.8 &50.7 &\multicolumn{1}{l|}{67.4} &9.6 &6.6 &4.7 &5.1 &6.5 \\
\multicolumn{1}{l|}{}           & \multicolumn{1}{l|}{FDA \cite{yang2020fda}}              &12.5 &0.0  &\multicolumn{1}{l|}{6.3}  &13.5 &20.5 &\multicolumn{1}{l|}{17.0} &46.5 &68.4 &73.8 &55.7 &\multicolumn{1}{l|}{61.1} &10.2 &7.6 &4.3 &5.0 &6.8 \\ \hline

\multicolumn{1}{l|}{\multirow{3}{*}{\rotatebox[origin=c]{90}{\scriptsize Medical}}} & \multicolumn{1}{l|}{SIFA \cite{chen2020unsupervised}}             &47.2 &19.7 &\multicolumn{1}{l|}{33.5} &19.1 &3.3  &\multicolumn{1}{l|}{11.2} &76.3 &88.1 &74.2 &62.4 &\multicolumn{1}{l|}{75.3} &6.1 &3.9 &7.1 &4.8 &5.5 \\
\multicolumn{1}{l|}{}   &\multicolumn{1}{l|}{PSIGAN \cite{jiang2020psigan}}           &57.3 &37.2 &\multicolumn{1}{l|}{47.3} &5.5  &2.8  &\multicolumn{1}{l|}{4.1}  &67.2 &87.0 &80.3 &60.8 &\multicolumn{1}{l|}{73.8} &7.5 &4.2 &5.5 &\textbf{4.7} &5.5 \\
\multicolumn{1}{l|}{}   &\multicolumn{1}{l|}{SDC-UDA (ours)}          &\textbf{84.6} &\textbf{84.9} &\multicolumn{1}{l|}{\textbf{84.8}} &\textbf{0.51} &\textbf{0.14} & \multicolumn{1}{l|}{\textbf{0.33}} &\textbf{95.8} &\textbf{91.0} &\textbf{88.6} &\textbf{66.6} & \multicolumn{1}{l|}{\textbf{85.5}} &\textbf{1.1} &\textbf{2.8} &\textbf{3.2} &6.3 &\textbf{3.3} \\

\end{tabular}}
\begin{tablenotes}
  \item[1] Since the ground truth for CrossMoDA Dataset is not available, fully-supervised results are referred from the dataset owner's paper\cite{DORENT2023102628}. These figures correspond to the upper bound and are the evaluation results from larger test set (N=137) than the one used in our study.
\end{tablenotes}
\end{threeparttable}
\end{adjustbox}

  \label{tab:table31}

\end{table*}

\begin{figure*}[h]
\captionsetup{width=1.0\linewidth}
\centering\includegraphics[width=1\linewidth]{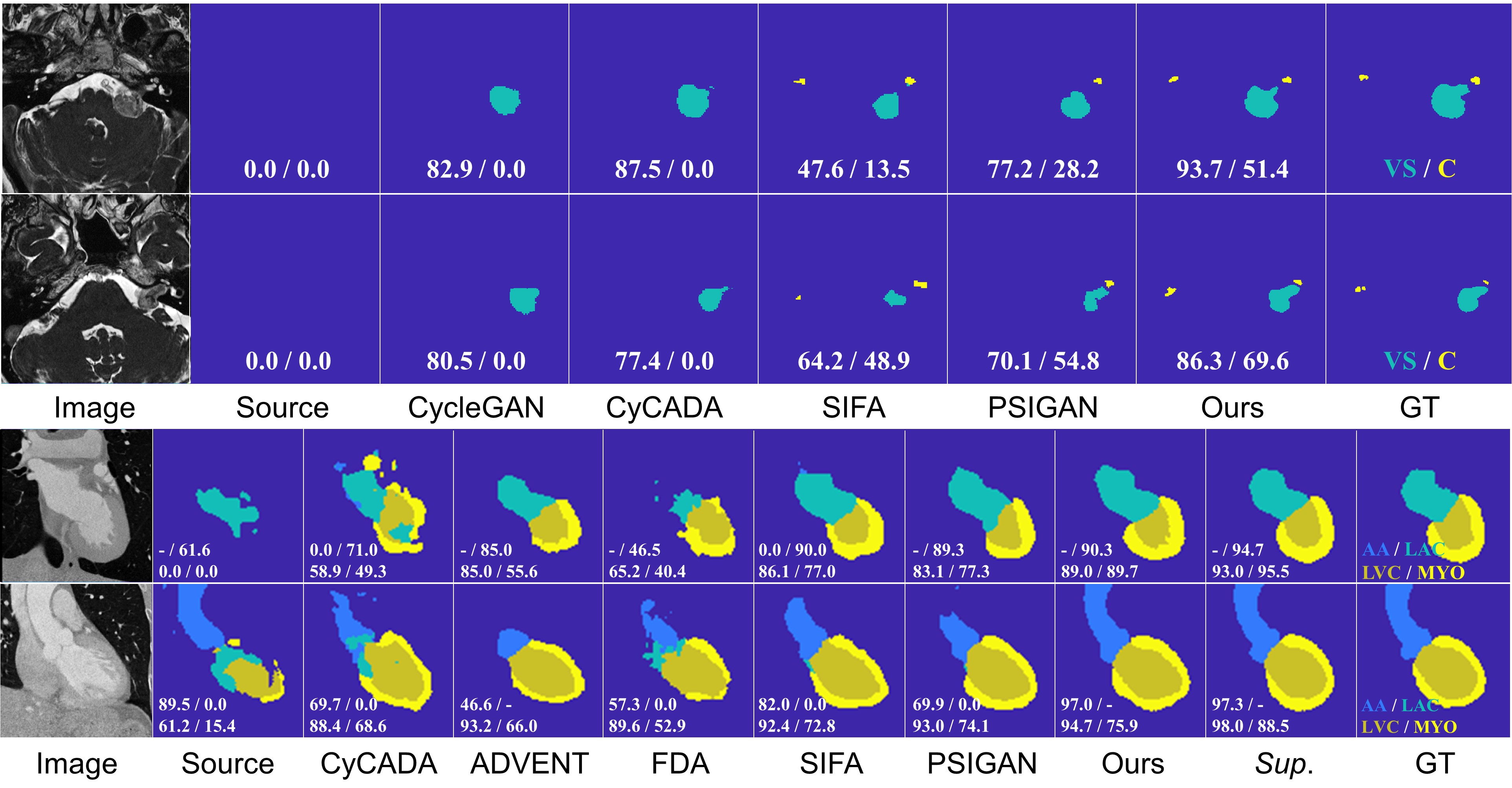}
\caption{Comparison of qualitative results between SDC-UDA and previous non-medical and medical UDA methods. Since the ground truth of CrossMoDA dataset is not accessible, we requested to the data owner a few of the label slices and used them to make this figure. The numbers indicate Dice values of each substructure in the slice. \textit{Sup} denotes the fully-supervised result.}
\label{fig:fig6}
\end{figure*}

\noindent \textbf{Volumetric self-training. } 
For volumetric segmentation self-training, 3D U-Net based architecture constructed from nnU-Net\cite{nnunet} was used. Segmentation was trained for 100 epochs using a combination of Dice and cross entropy loss with stochastic gradient descent optimizer. The initial learning rate and batch size were 1e-2 and 2, respectively. Detailed information on network architecture and training strategy can be found in the supplementary materials.

\subsection{Results}
\noindent \textbf{Effect of intra- and inter-slice self-attention module in image translation.}
\cref{fig:fig5} shows a representative case that presents how the quality of the synthetic image changes depending on the design of the image translation network. Compared with the proposed method (ours), slice-direction inconsistency of pixel intensity was observed in the translated image due to the variation derived from slice-by-slice prediction with a 2D network (blue arrows). This was not completely overcome even with a 2.5D network that puts two adjacent slices together in the network. Also, in both 2D and 2.5D networks, it was found that the anatomy of the source image was distorted in the translated image (red and green arrows). On the other hand, it was observed that the aforementioned problems were alleviated when utilizing the intra- and inter-slice self-attention module that effectively considers neighboring anatomy. \Cref{tab:table1} shows the increase of final segmentation performance by incorporating intra- and inter-slice self-attention in image translation network compared to baseline that utilizes pure unpaired image translation followed by 3D segmentation network.\\

\begin{figure}[t]
\captionsetup{width=1.0\linewidth}
\centering\includegraphics[width=1\linewidth]{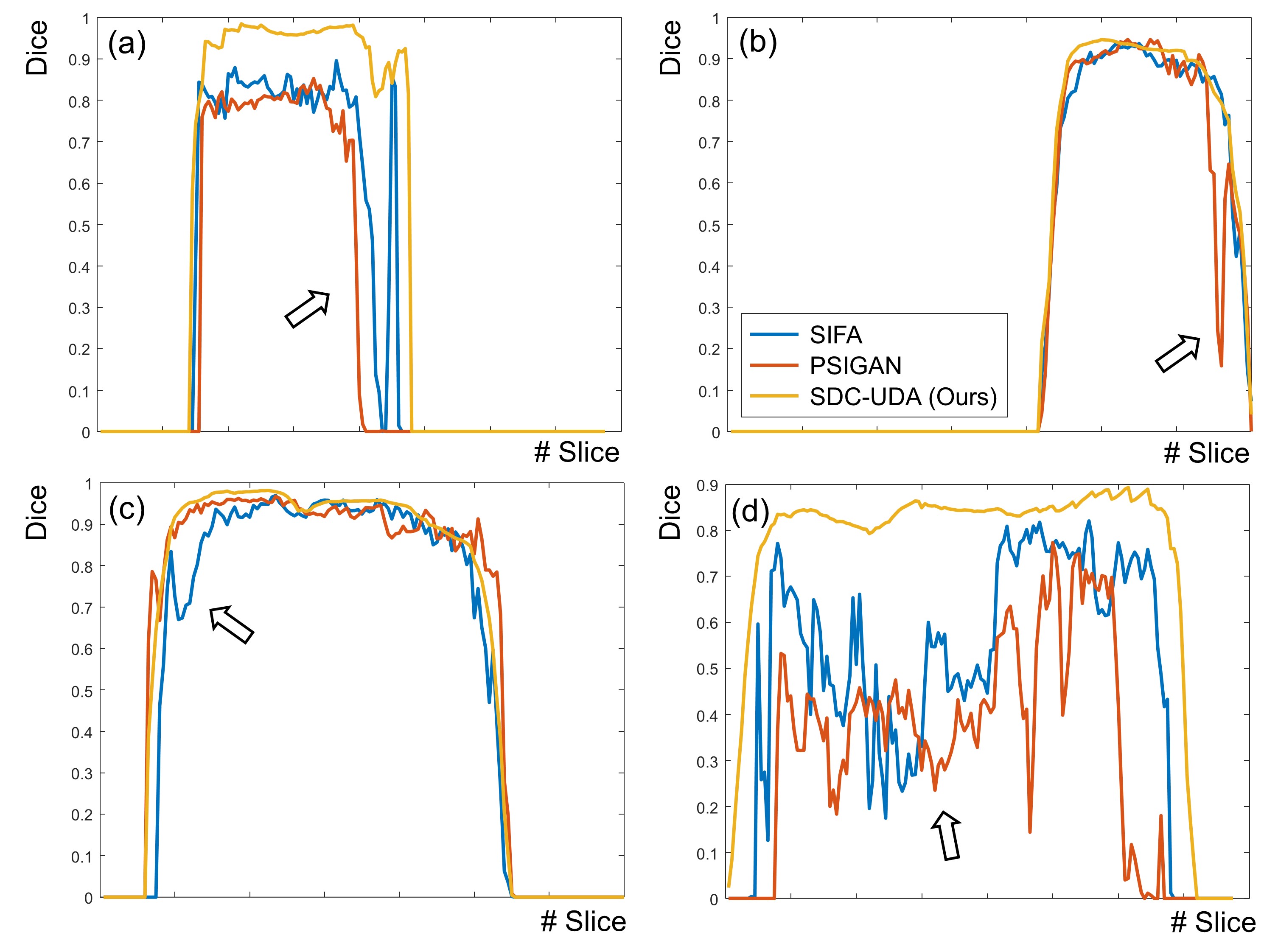}
\caption{Representative cases demonstrating the superior slice-direction continuity of segmentation by the proposed method compared to other medical UDA methods on cardiac dataset. The plots show graphs of Dice coefficient for each slice. (a), (b), (c), and (d) denote AA, LAC, LVC, and MYO class, respectively. Best viewed in color. }
\label{fig:fig7}
\end{figure}

\noindent \textbf{Effect of uncertainty-constrained pseudo-label refinement and volumetric self-training.} 
Since the target domain ground truth is inaccessible for the CrossMoDA dataset, the effectiveness of pseudo-label refinement was verified by comparing the performance of self-training before and after pseudo-label refinement. For cardiac datasets for which ground truth is available, it was verified by comparing dice coefficients with labels before and after pseudo-label refinement (please refer to supplementary material). \Cref{tab:table1} shows the improvement of target domain segmentation performance by applying the proposed uncertainty-constrained pseudo-label refinement on CrossMoDA dataset. It can be seen that volumetric self-training combined with the proposed uncertainty-constrained pseudo-label refinement increases the quantitative metrics in both VS and C by a large margin, whereas pure self-training only increase the performance in VS.  \\

\noindent \textbf{Comparative studies.}
The proposed method is compared with six popular UDA methods that are CycleGAN \cite{zhu2017unpaired}, CyCADA \cite{hoffman2018cycada}, ADVENT \cite{vu2019advent}, FDA \cite{yang2020fda}, SIFA \cite{chen2020unsupervised}, and PSIGAN \cite{jiang2020psigan}. The first four methods are from natural image field whereas the last two are UDA for medical image segmentation methods. Fully-supervised and without-adaptation results are presented to provide upper bound and lower bound, respectively. Please note that the full-supervised result of CrossMoDA dataset, for which the ground truth segmentation mask is not available, is referred from the paper of the dataset owner \cite{DORENT2023102628}. \\ \indent \Cref{tab:table31} and \cref{fig:fig6} shows quantitative and qualitative results of comparative studies between the proposed method and previous methods. SDC-UDA far exceeds the results of UDA for semantic segmentation studies in the non-medical field, and shows better results in multiple tasks compared to the recent studies in the medical field. In particular, there was a significant performance gap between the proposed method and comparison methods in the CrossMoDA dataset compared to the cardiac dataset. This is probably attributed to the fact that previous studies on medical UDA are dedicated to datasets in which the data are cropped around the target organs and the foreground objects occupy a large portion of the data. In contrast, SDC-UDA shows that both target structures in the CrossMoDA dataset are segmented with high accuracy.\\ 

\noindent \textbf{Slice-direction continuity of segmentation compared to previous studies.}
\cref{fig:fig7} shows the excellent slice-direction continuity of segmentation by the proposed method compared to the previous studies on the MMWHS cardiac dataset. Fig. \ref{fig:fig7}-(a), (b), (c), and (d) plot the slice-wise Dice values from representative cases of AA, LAC, LVC, and MYO, respectively. Our SDC-UDA shows gradual and consistent segmentation performance in the slice-direction whereas SIFA and PSIGAN, which are recently proposed medical UDA methods, show very inconsistent and sometimes sudden fluctuations of segmentation performance in the slice direction. This suggests the potential of the proposed method to be useful in the clinical practice where precise volumetric segmentation is required to analyze the patient's status with high confidence.


\section{Conclusion}
\label{sec:exp}

In this study, we proposed SDC-UDA, a novel volumetric UDA framework for slice-direction continuous cross-modality medical image segmentation, and validated it on multiple public datasets. SDC-UDA effectively translates medical volumes through intra- and inter-slice self-attention and better adapts to target domain via volumetric self-training enhanced by simple yet effective pseudo-label refinement strategy that utilizes uncertainty maps. Ablation studies demonstrated the effectiveness of each component and comparative studies showed the superior performance of the proposed method. \\

\noindent \textbf{Acknowledgements.} This research was supported by Samsung Research Funding Center of Samsung Electronics under Project Number SRFC-TF2103-01 in constructing hardware systems and multi-modal studies. It was also supported by Basic Science Research Program through the National Research Foundation of Korea funded by the Ministry of Science and ICT (2021R1A4A1031437, 2022R1A2C2008983), Artificial Intelligence Graduate School Program at Yonsei University [No. 2020-0-01361], the KIST Institutional Program (Project No.2E31051-21-204), and partially supported by the Yonsei Signature Research Cluster Program of 2022 (2022-22-0002).

{\small
\bibliographystyle{ieee_fullname}
\bibliography{SDCUDA_arXiv}
}

\end{document}